\documentclass[10pt,twocolumn,letterpaper]{article}

\usepackage{cvpr}              

\definecolor{cvprblue}{rgb}{0.21,0.49,0.74}
\usepackage[pagebackref,breaklinks,colorlinks,allcolors=cvprblue]{hyperref}

\def\paperID{CSC2530H} 
\def\confName{CSC2530H}
\def\confYear{2025}

\title{Improving Open-Set Semantic Segmentation in 3D Point Clouds\\ by Conditional Channel Capacity Maximization:\\ Preliminary Results}

\author{Wang Fang; Shirin Rahimi; Olivia Bennett; Sophie Carter; Mitra Hassani; Xu Lan; Omid Javadi; Lucas Mitchell\\
University of Toronto\\
27 King's College Cir \\
}

\begin{document}
\maketitle

\begin{abstract}
Point‑cloud semantic segmentation underpins a wide range of critical applications. Although recent deep architectures and large‑scale datasets have driven impressive closed‑set performance, these models struggle to recognize or properly segment objects outside their training classes. This gap has sparked interest in Open‑Set Semantic Segmentation (O3S), where models must both correctly label known categories and detect novel, unseen classes. In this paper, we propose a plug and play framework for O3S. By modeling the segmentation pipeline as a conditional Markov chain, we derive a novel regularizer term dubbed Conditional Channel Capacity Maximization (3CM), that maximizes the mutual information between features and predictions conditioned on each class. When incorporated into standard loss functions, 3CM encourages the encoder to retain richer, label‑dependent features, thereby enhancing the network’s ability to distinguish and segment previously unseen categories. Experimental results demonstrate effectiveness of proposed method on detecting unseen objects. We further outline future directions for dynamic open‑world adaptation and efficient information‑theoretic estimation. 
\end{abstract}

\section{Introduction}
\label{sec:intro}

Point‑cloud semantic segmentation has attracted intense interest from both academia and industry, driven by the proliferation of 3D sensing technologies such as LiDAR and structured‑light scanners. Its applications span critical domains—autonomous driving \cite{9000872, Chen_2016_CVPR}, service and industrial robotics \cite{9636711, 6696655, 4399309}, and medical imaging for tasks like anatomical structure delineation \cite{9127813}—where dense, per‑point labeling enables downstream functions such as path planning, object manipulation, and surgical navigation. Beyond enabling richer scene understanding, accurate point‑wise segmentation reduces reliance on expensive manual annotation and facilitates end‑to‑end learning pipelines for tasks like object detection, mapping, and human‑robot interaction.
Recent years have witnessed remarkable breakthroughs, powered by the advent of deep neural architectures tailored to unstructured point sets. Early models such as PointNet \cite{qi2017pointnet} demonstrated that simple, permutation‑invariant networks could directly consume raw point clouds, while follow‑ups like KPConv \cite{thomas2019kpconv} and MinkowskiNet \cite{choy20194d} introduced flexible, convolutional operations on irregular 3D data. These architectures, coupled with large‑scale benchmarks (e.g., ShapeNet Part \cite{wu20153d}) and enhanced computational infrastructure (multi‑GPU, distributed training), now routinely achieve state‑of‑the‑art accuracy under the closed‑set protocol—where training and test classes coincide.

In practical deployments, however, this closed‑set assumption breaks down. Real‑world scenes often contain objects and categories unseen during training, leading standard models to misclassify novel instances or assign overconfident—but incorrect—labels. The resulting performance degradation motivates the study of Open‑Set Semantic Segmentation (O3S), which seeks robust methods capable of both recognizing known categories and detecting or adapting to previously unseen classes. Existing O3S strategies—ranging from uncertainty estimation and prototype‑based separation \cite{10655771, 10204183} to contrastive clustering and synthetic unknown generation—have made important strides, yet often rely on task‑specific heuristics or complex generative components.

To address these limitations, we propose Conditional Channel Capacity Maximization (3CM), a theoretically grounded regularization term that can be seamlessly integrated into any point‑cloud segmentation framework. Drawing on an information‑theoretic model of segmentation as a conditional Markov chain, 3CM encourages the feature encoder to preserve richer, class‑conditional information—thereby enhancing the model’s ability to distinguish and adapt to unseen categories during inference. Importantly, this plug‑and‑play term requires neither architectural modification nor additional generative modules.

Through extensive experiments on multiple open‑set benchmarks, we demonstrate that augmenting standard segmentation losses with 3CM consistently improves both known‑class accuracy and novel‑class detection, outperforming prior O3S methods while retaining high efficiency. 

 In summary, Our contributions are threefold:
 
\begin{itemize}
    \item We study the problem of open-set semantic segmentation through the lens of information theory by modeling it as a conditional Markov chain.
    \item A novel regularization term, dubbed Conditional Channel Capacity Maximization (3CM), is proposed based on the information theory model we constructed.
    \item By integrating 3CM with existing 3D point cloud segmentation algorithms, the model's open-set semantic segmentation performance consistently improves, illustrating the effectiveness of the proposed method.
\end{itemize}
\section{Related Work}
\label{sec:RelatedWork}
In this section, we briefly review the literature on open set semantic segmentation in 3D point clouds as a critical task in computer vision. Despite advances in deep learning architectures for 3D point clouds (e.g., PointNet \cite{qi2017pointnet}, KPConv \cite{thomas2019kpconv}, MinkowskiNet \cite{choy20194d}), the irregular, sparse, and unstructured nature of point cloud data makes identifying unknown instances particularly difficult. 

Recent studies address these challenges through a diverse set of approaches, including uncertainty estimation, prototype-based learning, contrastive clustering, and synthetic unknown generation. For instance, \citet{10655771} introduce a Probability-Driven Framework (PDF) for 3D open-set segmentation, incorporating uncertainty estimation and pseudo-labeling to detect unknown classes while enabling incremental adaptation to newly identified objects. Similarly, \citet{10204183} propose an Adversarial Prototype Framework (APF), which learns class prototypes in feature space while employing generative models to estimate unseen-class distributions, facilitating the separation of known and unknown points. These methods illustrate growing recognition that latent space representations and probabilistic modeling are key to tackling open-world segmentation.

Another major research trend involves clustering-based and contrastive-learning methods for segmenting unknown classes, \citet{10203892} leverage batch-level clustering and pseudo-labeling to iteratively refine the identification of novel classes, ensuring robustness in complex 3D scenes like SemanticKITTI. Furthermore, \citet{Hong2022PointCaMCF} introduce PointCaM, a compelling strategy that generates synthetic unknown examples by geometrically manipulating known data during training, highlighting the importance of data augmentation in modeling unknown distributions. More recent works such as AutoVoc3D \cite{wei2024autovocabularysegmentationlidarpoints} take an innovative step by integrating open-vocabulary segmentation with a text-point semantic similarity (TPSS) metric, suggesting that multi-modal approaches may enrich the robustness of OSS in 3D environments.

Differ from all previous works, in this paper, we study the open-set semantic segmentation task through the lens of information theory. 
\section{Notation}

For a positive integer $C$,  let $[C]\triangleq \{1,\dots,C\}$. Denote by $P[i]$ the $i$-th element of vector $P$.  
For two vectors $U$ and $ V$, denote by $\langle U, V\rangle$ their inner product. For two matrices $M\in \mathbb{R}^{m\times n}$ and $N\in \mathbb{R}^{n\times k}$, denote by $M@N$ their matrix product. We use $|\mathcal{C}|$ to denote the cardinality of a set $\mathcal{C}$. 
The entropy of $C$-dimensional probability vector $P$ is defined as $\mathrm{H} (P) = \sum_{c=1}^C -P [c] \log P [c]$. Also, the cross entropy of two $C$-dimensional probability vectors $P_1$ and $P_2$ is defined as $\mathrm{H} (P_1, P_2) = \sum_{c=1}^C -P_1 [c] \log P_2 [c]$, and their Kullback–Leibler ($\mathrm{KL}$) divergence is defined as $\mathrm{KL} (P_1 || P_2 ) = \sum_{c=1}^C P_1 [c] \log {\frac{P_1 [c]}{P_2 [c]}}$.

For a random variable $X$, denote by $P_{X}$ its probability distribution, and by $E_{X}[\cdot]$ the expected value w.r.t. $X$. For two random variables $X$ and $Y$, denote by $\mathbb{P}_{(X,Y)}$ their joint distribution. The mutual information between two random variables $X$ and $Y$ is written as $\mathrm{I}(X;Y)$, and the conditional mutual information of $X$ and $Y$ given a third random variable $Z$ is $\mathrm{I}(X;Y|Z)$. Consider a dataset $\mathcal{D}$ of size $n$ with $C$ classes, $\mathcal{D} = \{(\boldsymbol{x}_i, y_i)\}_{i=1}^n$, where each $\boldsymbol{x}_i \in \mathbb{R}^d$ and $y_i \in [C]$. For any class $y$, we define $\mathcal{D}_y = \{(\boldsymbol{x}_j,y_j) \in \mathcal{D} | y_j = y\}$ the subset of $\mathcal{D}$ containing all samples with label $y$. Lastly, the softmax operation is denoted by $\sigma(\cdot)$. 

\section{Methodology}
\subsection{Segmentation as a Markov Chain}


In this project, we primarily use PointNet \cite{qi2017pointnet} as the feature extractor. PointNet takes $n$ points as input, applies input and feature transformations, and aggregates point-wise features through max pooling. The output is a set of classification scores across $k$ classes. For segmentation, the network builds upon the PointNet backbone. Instead of aggregating all feature vectors globally, the segmentation head employs interpolation and a unit PointNet to produce per-point predictions \cite{10.5555/3295222.3295263}.

Put the neural network architecture aside, we can model the segmentation process as a Markov chain \cite{10900607, yang2025methods}. Each point $X$, labeled with $Y = y$ and accompanied by its neighbors, can be viewed as a sample from class $y$. A deep neural network (PointNet, in our case) maps this sample to a feature representation $\hat{X}$, which is then further processed by the segmentation head to yield an output distribution $\hat{Y}$. Notably, the sample $X$ is drawn from the conditional distribution $P(X|Y)$.

Modeling segmentation as a Markov chain allows us to analyze it using information-theoretic tools \cite{6773024, cover1999elements}. In particular, we introduce a quantity called the conditional mutual information between the prediction $\hat{Y}$ and the sample $X$, given the ground-truth label $Y = y$, following the definitions in \cite{10900607, ye2024bayes}.

\begin{align}
    &I(\hat{Y}, X| Y)\\
    = &\sum_x P_{X|Y}(x|y)\Bigg[\sum_{i=1}^CP(\hat{Y} = 1|x) \log\frac{P(\hat{Y} = i|x)}{P_{\hat{Y}|y}(\hat{Y} = i | Y=i)}\Bigg] \\
    = & \mathbb{E}_{X|Y}\Bigg[\Bigg( \sum_{i=1}^C P_X[i] \log \frac{P_X[i]}{P_{\hat{Y}|y}(\hat{Y}=i|Y=i)}\Bigg) | Y=y\Bigg] \\
    = & \mathbb{E}_{X|Y} \Big[ \mathrm{KL}(P_X\|P_{\hat{Y|y}})|Y=y\Big].\label{Eq:CMI}
\end{align}

Given the conditional mutual information of class $y$, on the other hand, the channel capacity of the model $f$ can be written as:


\begin{align} \label{eq:emp_CMI}
&\text{CMI}_{\rm emp}(f)= \frac{1}{N} \sum_{y \in [C]} \sum_{x_j \in \mathcal{D}_{y}} \text{KL} (P_{x_j} ~ || ~ Q_{\rm emp}^y), \\  \label{eq:Qemp}
\text{where}~~~~~~~ & Q_{\rm emp}^y= \frac{1}{|\mathcal{D}_{y}|} \sum_{x_j \in \mathcal{D}_{y}} P_{x_j}, ~~ \text{for} ~~ y \in [C]. 
\end{align}

\subsection{Necessity of Maximizing Conditional Channel Capacity}


As discussed in the previous section, given a dataset $\mathcal{D}$, the empirical conditional mutual information (CMI) can be quantified using 

. This measures how much information the model's output prediction retains about the input, conditioned on the ground-truth label \cite{hamidi2024adversarialtrainingadaptiveknowledge}.

In open-set semantic segmentation, a special class—often referred to as the unknown class—represents categories that are not seen during training \cite{10.1007/978-3-031-73024-5_10}. This class may contain samples from multiple distinct, unseen categories. We illustrate this concept using the Venn diagrams in Fig. 1.

Fig. 1(a) shows the information structure of a model trained solely with cross-entropy loss. In this setting, the model only preserves the minimal information required to distinguish the unknown class $Y_{unknown}$ from the known classes. However, the extracted features might not capture all relevant information about the underlying unseen categories.

In contrast, Fig. 1 depicts a scenario where we increase the CMI for the unknown cluster $Y_{unknown}$, encouraging the model to preserve richer information that may be useful for differentiating among all potential (yet unseen) classes within this cluster.

In the following section, we describe our approach to maximizing the conditional mutual information in this context.

\begin{table*}[th!]
\caption{Segmentation results on the ShapeNet Part dataset. The evaluation metric is mean IoU (mIoU) computed over points. We compare our model's performance on both known and unknown sets, with (w.) and without (w.o.) 3CM, against two traditional methods \cite{WU2014248, 10.1145/2980179.2980238} and a 3D fully convolutional network baseline proposed by us. Our PointNet-based method achieves state-of-the-art performance in mIoU.}
\label{Tab:SegResults}
\resizebox{1\textwidth}{!}{\begin{tabular}{ccccccccccc|ccccccccc}
\hline
         &                                                    & \multicolumn{9}{c|}{Known classes}                                                                               & \multicolumn{9}{c}{Unknown classes}                                                                                   \\
         & Training                                           & mean & areo & bag  & cap  & car  & chair & \begin{tabular}[c]{@{}c@{}}ear\\  phone\end{tabular} & guitar & knife & mean & lamp & laptop & motor & mug  & pistol & rocket & \begin{tabular}[c]{@{}c@{}}skate\\ board\end{tabular} & table \\
\# shape & Method                                             & 1091 & 2690 & 76   & 55   & 898  & 3758  & 69                                                   & 787    & 392   & 1020 & 1547 & 451    & 202   & 184  & 283    & 66     & 152                                                   & 5271  \\ \hline
Wu \cite{WU2014248}      & \begin{tabular}[c]{@{}c@{}}w.o\\  3CM\end{tabular} & -    & 70.2 & -    & -    & -    & 80.5  & -                                                    & -      & -     & -    & 53.7 & -      & -     & -    & -      & -      & -                                                     & 60.4  \\
         & \begin{tabular}[c]{@{}c@{}}w. \\ 3CM\end{tabular}  & -    & 70.6 & -    & -    & -    & 80.4  & -                                                    & -      & -     & -    & 76.5 & -      & -     & -    & -      & -      & -                                                     & 73.1  \\ \hline
Yi \cite{10.1145/2980179.2980238}   & \begin{tabular}[c]{@{}c@{}}w.o \\ 3CM\end{tabular} & 80.9 & 82.6 & 79.2 & 79.1 & 76.2 & 90.1  & 62.3                                                 & 90.5   & 87.4  & 58.8 & 75.2 & 43.2   & 60.7  & 85.8 & 72.3   & 28.7   & 43.8                                                  & 60.5  \\
         & \begin{tabular}[c]{@{}c@{}}w. \\ 3CM\end{tabular}  & 81.7 & 83.7 & 80.6 & 79.0 & 77.3 & 89.4  & 65.3                                                 & 89.4   & 89.2  & 75.6 & 81.6 & 94.7   & 71.3  & 89.2 & 82.7   & 48.2   & 65.7                                                  & 71.6  \\ \hline
3DCNN \cite{6165309}  & \begin{tabular}[c]{@{}c@{}}w.o \\ 3CM\end{tabular} & 77.6 & 77.5 & 72.4 & 76.3 & 72.8 & 87.2  & 65.1                                                 & 89.4   & 80.2  & 55.5 & 50.7 & 85.4   & 43.7  & 72.6 & 65.4   & 20.7   & 61.5                                                  & 43.7  \\
         & \begin{tabular}[c]{@{}c@{}}w. \\ 3CM\end{tabular}  & 77.5 & 77.4 & 73.4 & 76.3 & 71.7 & 86.4  & 64.7                                                 & 90.1   & 80.4  & 72.2 & 72.5 & 93.4   & 61.0  & 85.7 & 72.9   & 50.4   & 65.4                                                  & 76.0  \\ \hline
PointNet \cite{qi2017pointnet} & \begin{tabular}[c]{@{}c@{}}w.o\\ 3CM\end{tabular}  & 82.9 & 84.6 & 80.1 & 82.5 & 72.4 & 89.1  & 76.4                                                 & 92.8   & 85.5  & 64.4 & 60.3 & 87.5   & 47.6  & 81.8 & 64.7   & 33.1   & 68.4                                                  & 71.8  \\
         & \begin{tabular}[c]{@{}c@{}}w. \\ 3CM\end{tabular}  & 83.2 & 83.9 & 79.3 & 85.4 & 72.4 & 89.7  & 77.2                                                 & 91.4   & 86.0  & 77.0 & 80.1 & 92.4   & 65.1  & 90.3 & 80.4   & 55.9   & 72.3                                                  & 79.8  \\ \hline
\end{tabular}}
\end{table*}

\subsection{Conditional Channel Capacity Maximization}

As outlined in the previous section, enabling the model to effectively classify unseen classes requires it to retain sufficient information in its feature representations. To encourage this, we introduce a regularization term that promotes information preservation within the learned features \cite{ye2024bayes}. Accordingly, we incorporate the following objective into the training process and optimize it to maximize the model's capacity for representing unknown classes:

\begin{align}
    \mathcal{L}_{3CM} = \frac{1}{N_{unknown}} \sum_{x_j\in D_{unknown}} \text{KL}(P_{x_j}\|\hat{Q}^y_{emp}), \label{Eq:ObjectiveFn}
\end{align}

where $\hat{Q}^y_{emp} =  \frac{1}{N_{unknown}} \sum_{x_j\in D_{unknown}} P_{x_j}$. 

During training, whenever a sample belonging to the unknown class is present in the current batch \cite{hamidi2025distributed}, the additional objective defined in Eq. \ref{Eq:ObjectiveFn} is activated. This term is designed to enhance the model’s ability to preserve information relevant for distinguishing unseen classes \cite{10446776}. A non-negative hyperparameter $\lambda$ is introduced to control the influence of this regularization term, allowing us to balance the trade-off between standard segmentation performance and the information preservation objective. The modified loss function is defined as follows\footnote{If the impact of random mini-batch sampling and SGD is ignored, the alternating algorithm is guaranteed to converge in theory since given $\theta$, the optimal $\hat{Q}^y_{emp}$ can be found analytically via $\hat{Q}^y_{emp} =  \frac{1}{N_{unknown}} \sum_{x_j\in D_{unknown}} P_{x_j}$, although it might not converge to the global minimum. In the Ablation Study Section \ref{Sec:Ablation3CM}.}:

\begin{align}
    \min_\theta\ \ &\text{CE}(P_{x_{unknown}}, Y_{unknown}) + \nonumber \\
    &\lambda \mathcal{L}_{3CM} (P_{x_{unknown}},\hat{Q}^y_{emp}), \label{Eq:totalObject}
\end{align}

then we update $\hat{Q}^y_{emp}$ using an exponential moving average (EMA) to make updating stable.
\begin{align}
    \hat{Q}^y_{emp}\leftarrow\beta\hat{Q}^y_{emp}+(1-\beta)P_{x_{unknown}}
\end{align}
Here, $\beta$ is the EMA (Exponential Moving Average) factor, a predefined hyperparameter, which we set to 0.995 in all our experiments\footnote{In the Ablation Study Section \ref{Sec:AblationEMA} we study the impact of the EMA factor.}.
In the next section, we present a comprehensive set of experiments to verify the effectiveness of the proposed method.

\section{Experimental Results}
In this section, we conclude the paper by conduct a through set of experiments. To this end, we follow the original PointNet \cite{qi2017pointnet} settings, which contains following two applications:

\begin{table}[h!]
\caption{Classification results on ModelNet40 \cite{ye2025towards}. Compared to baseline methods, 3CM consistently improves accuracy on unseen classes.}
\label{Tab:ClassifierRes}
\resizebox{\linewidth}{!}{\begin{tabular}{c|cccc}
\hline
            & method                                              & \#views & \begin{tabular}[c]{@{}c@{}}accuracy\\ seen classes\end{tabular} & \begin{tabular}[c]{@{}c@{}}accuracy\\ unseen classes\end{tabular} \\ \hline
3DShapeNets & \begin{tabular}[c]{@{}c@{}}w.o.\\  3CM\end{tabular} & 1       & 79.4                                                            & 56.8                                                              \\
            & \begin{tabular}[c]{@{}c@{}}w. \\ 3CM\end{tabular}   & 1       & 78.1                                                            & 72.5                                                              \\
Subvolume   & \begin{tabular}[c]{@{}c@{}}w.o. \\ 3CM\end{tabular} & 20      & 87.1                                                            & 60.7                                                              \\
            & \begin{tabular}[c]{@{}c@{}}w. \\ 3CM\end{tabular}   & 20      & 87.6                                                            & 81.8                                                              \\
MVCNN       & \begin{tabular}[c]{@{}c@{}}w.o. \\ 3CM\end{tabular} & 80      & 95.4                                                            & 70.1                                                              \\
            & \begin{tabular}[c]{@{}c@{}}w.\\  3CM\end{tabular}   & 80      & 96.1                                                            & 90.0                                                              \\
PointNet    & \begin{tabular}[c]{@{}c@{}}w.o. \\ 3CM\end{tabular} & 1       & 88.1                                                            & 61.3                                                              \\
            & \begin{tabular}[c]{@{}c@{}}w.\\  3CM\end{tabular}   & 1       & 86.3                                                            & 85.9                                                              \\ \hline
\end{tabular}}
\end{table}

\begin{figure}
  \centering
   \includegraphics[width=\linewidth]{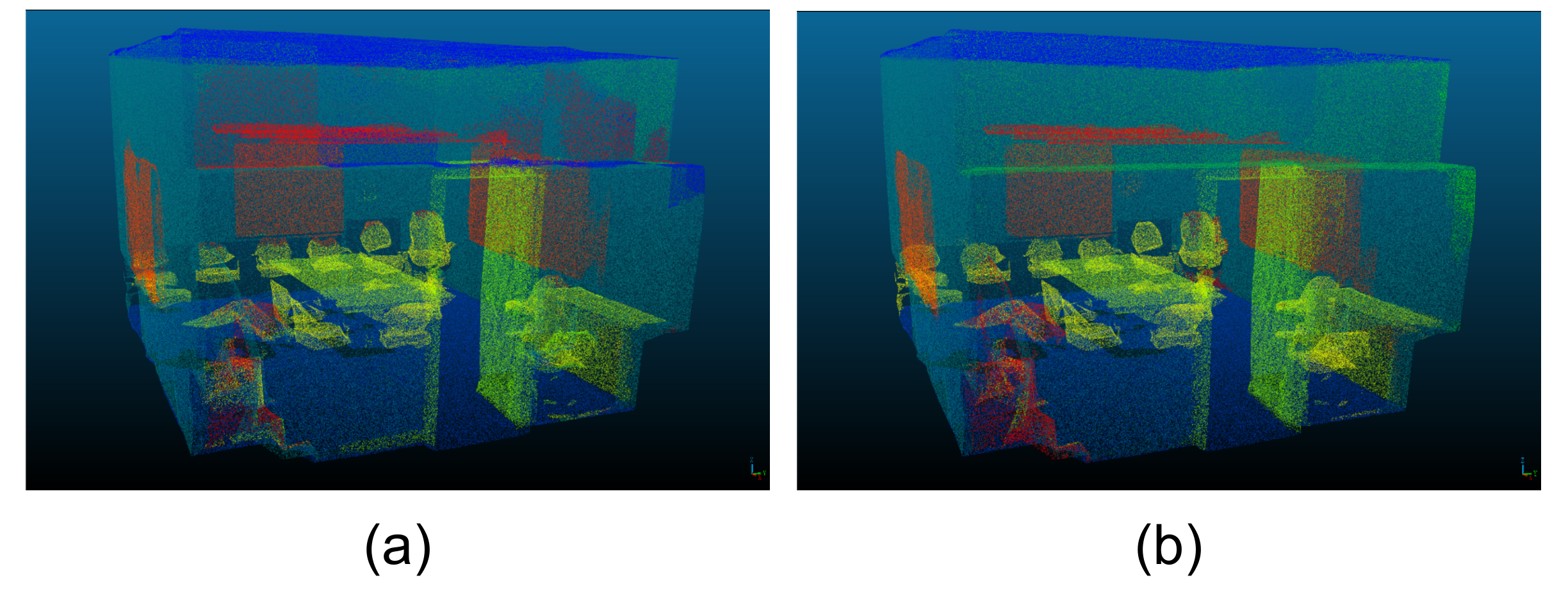}
   \caption{Semantic segmentation results on Stanford-indoor3d dataset with (b) and without (a) 3CM regularization. The fine-tuned model demonstrates significantly improved performance on unseen classes when 3CM is applied, which aligns with the results in Table \ref{Tab:SegResults}, confirming the effectiveness of 3CM.} \label{fig:segresults}
\end{figure}

\noindent $\bullet$ \textbf{3D Object Part Segmentation} Part segmentation is a fine-grained 3D recognition task that assigns a part category label (e.g., chair leg, cup handle) to each point or face of a 3D scan or mesh. We evaluate on the ShapeNet Part dataset [29], which includes 16,881 shapes across 16 object categories, annotated with a total of 50 part types. Most categories are annotated with 2–5 parts, with labels provided on sampled points \cite{ye2025towards}.

 We group the classes lamp, laptop, motor, mug, pistol, rocket, skateboard, and table into a single "unknown" class. After the initial training phase, we remove the output column corresponding to the unknown class, append 8 new output columns, and train the model for an additional 30 epochs.

We formulate the task as a per-point classification problem and evaluate using mean Intersection over Union (mIoU). For a shape $S$ in category $C$, we compute IoU for each part type by comparing predicted and ground-truth points \cite{10900607}. If both are empty for a part, the IoU is set to 1. The shape-level mIoU is the average over all part types in category $C$, and the category-level mIoU is the mean over all shapes in that category.

In this section, we evaluate our method on PointNet with and without the proposed 3CM regularization term. We observe that incorporating 3CM leads to a significant improvement in opem-set segmentation performance, particularly in distinguishing between previously unknown part classes.

As shown in Table \ref{Tab:SegResults}, on the ShapeNet Part dataset, applying the 3CM regularization term to unseen classes followed by post-training consistently improves model performance \cite{10619204}. The proposed 3CM method notably outperforms the baseline, with the most significant improvements observed on the unseen classes.

\begin{table}[th!]
\centering
\caption{ Semantic segmentation results (IoU) on the Stanford-Indoor3D dataset: without 3CM regularization and with 3CM regularization.}
\label{Tab:Standfordres}
\resizebox{\linewidth}{!}{\begin{tabular}{ccc}
\hline
IoU      & Seen Classes & Unseen Classes \\ \hline
w.o. 3CM & 85.4         & 62.3           \\
w. 3CM   & 85.3         & 80.1           \\ \hline
\end{tabular}}
\end{table}

We further evaluate our method by integrating PointNet on the Stanford Indoor3D dataset \cite{10619241}. The key results are summarized in Table \ref{Tab:Standfordres}, while Figure \ref{fig:segresults} visualizes the semantic segmentation outputs. As shown in Figure \ref{fig:segresults}, models without 3CM regularization (a) exhibit significantly weaker performance on unseen classes compared to those with 3CM (b). This aligns with Table \ref{Tab:Standfordres}, where 3CM achieves comparable accuracy on seen classes (85.3 vs. 85.4 without 3CM) and delivers a +17.8 point improvement on unseen classes (80.1 vs. 62.3). These findings, consistent with Table \ref{Tab:SegResults}, underscore 3CM’s effectiveness in enhancing generalization to unseen classes.

\noindent $\bullet$ \textbf{3D Object Classification} We evaluate the effectiveness of our proposed method on the widely used ModelNet40 benchmark dataset \cite{wu20153d}, which consists of 12,311 CAD models from 40 categories of man-made objects. The dataset is split into 9,843 models for training and 2,468 for testing. Unlike prior approaches that rely on volumetric grids or multi-view image projections, our method operates directly on raw point cloud data, enabling more efficient and flexible 3D representation learning.

We uniformly sample 1,024 points from mesh surfaces, weighted by face area, and normalize each shape to fit within a unit sphere. During training, we apply standard data augmentation techniques, including random rotation around the up-axis and point-wise jittering with Gaussian noise (mean 0, standard deviation 0.02), to enhance generalization to unseen shapes and orientations.

As shown in Table \ref{Tab:ClassifierRes}, we compare our approach with both baselines used in the PointNet paper and state-of-the-art models.

Notably, when combine our regularization term with existing methods, their performance consistently improved in most cases \cite{yang2309conditional}.

\begin{figure}
  \centering
   \includegraphics[width=0.8\linewidth]{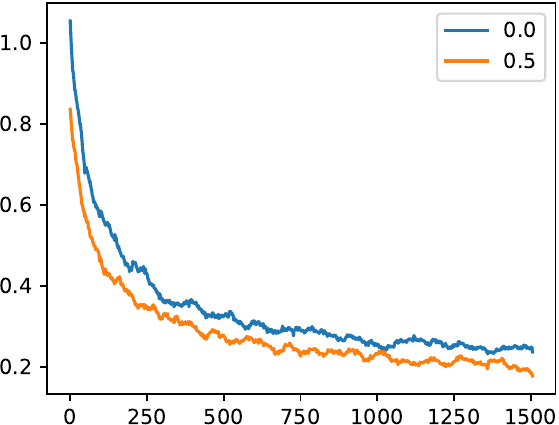}
   \caption{The objective value over the unseen class, the value converge with respect to the number of the training steps.} \label{fig:Converg}
\end{figure}
\section{Ablation Study}
\subsection{Effectiveness of the 3CM}\label{Sec:Ablation3CM}
\begin{table}[h!]
\centering
\caption{Ablation study on the $\lambda$ value. The optimal performance is achieved at $\lambda = 0.5$. Larger values lead to non-convergence due to unstable gradients.}
\label{Tab:AblationLambda}
\resizebox{\linewidth}{!}{\begin{tabular}{c|cccc}
\hline
IoU            & 0.1  & 0.3  & 0.5  & 0.7 \\ \hline
Seen Classes   & 85.4 & 85.6 & 85.3 & 72.1  \\
Unseen Classes & 74.1 & 77.9 & 80.1 & 50.4  \\ \hline
\end{tabular}}
\end{table}
In this section, we evaluate the effectiveness of the proposed 3CM regularization term by varying the weighting factor $\lambda$ in the objective function (Eq. \ref{Eq:totalObject}) within the range $[0, 0.7]$. Figure \ref{fig:Converg} shows the training loss curves for two representative cases: $\lambda = 0.0$ and $\lambda = 0.5$. In both cases, the loss converges, which supports the convergence analysis presented earlier.

\begin{figure}[!t]
  \centering
   \includegraphics[width=\linewidth]{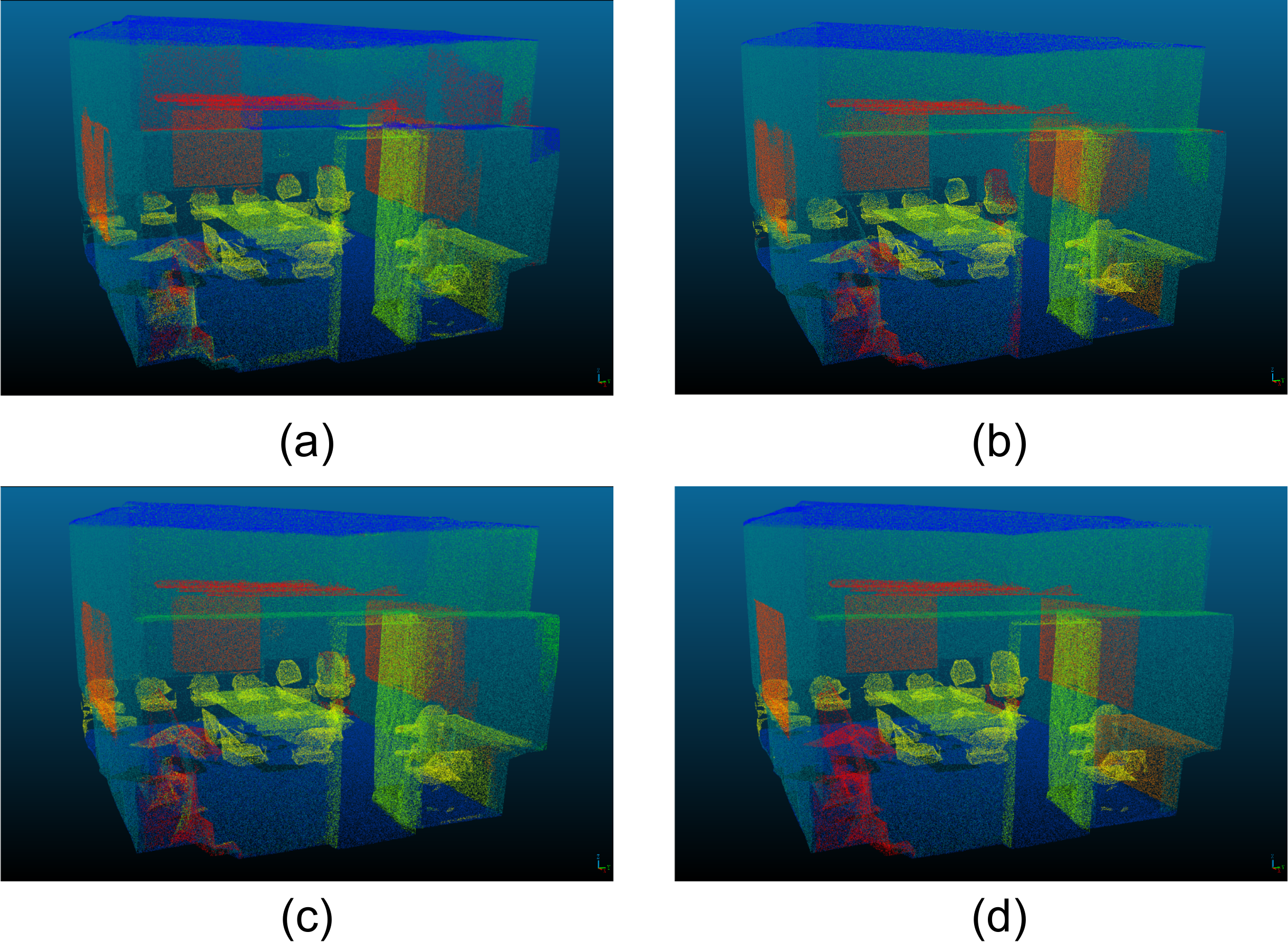}
   \caption{Segmentation results using varying regularization parameter  $\lambda$: (a)  $\lambda$=0.1, (b)  $\lambda$=0.3, (c)  $\lambda$=0.5, and (d)  $\lambda$=0.7.} \label{fig:SegFigDiffLambda}
\end{figure}

To further investigate the impact of $\lambda$, we report the model's performance on the Stanford-Indoor3D dataset under different $\lambda$ settings in Table \ref{Tab:AblationLambda}. We observe that the best performance is achieved when $\lambda = 0.5$. Larger values lead to unstable gradients and hinder convergence. Fig. \ref{fig:SegFigDiffLambda} presents segmentation results from models trained with varying $\lambda$ values. The model demonstrates consistent performance across the $\lambda$ range of $[0.1 - 0.5]$, but exhibits significant degradation at higher values, aligning with the quantitative trends observed in Tab. \ref{Tab:AblationLambda}.

\subsection{The necessary of using EMA to update the Q}\label{Sec:AblationEMA}

Table \ref{Tab:AblationEMA} presents the performance of PointNet under different EMA factors. The model performs reasonably well when the EMA factor is in the range of 0.99 to 0.999. However, when the factor is too small, the accuracy drops significantly. In the extreme case of $\beta = 0$, the model fails to converge.

\begin{table}[h!]
\centering
\caption{Ablation study on the EMA factor. The optimal performance is achieved at $\beta = 0.995$.}
\label{Tab:AblationEMA}
\resizebox{\linewidth}{!}{\begin{tabular}{c|ccccl}
\hline
EMA Factor     & 0  & 0.9  & 0.99 & 0.995 & 0.999 \\ \hline
Seen Classes   & NA & 81.6 & 86.4 & 85.3  & 84.2  \\
Unseen Classes & NA & 76.9 & 79.2 & 80.1  & 78.1  \\ \hline
\end{tabular}}
\end{table}

\section{Discussion}

While our approach demonstrates improved capability in preserving information relevant to open-set segmentation, it also has several limitations that open directions for future research. First, our method relies on estimating conditional mutual information (CMI), which can be computationally intensive and potentially sensitive to estimation noise, especially in high-dimensional feature spaces. Developing more robust and efficient estimators or incorporating approximation techniques could enhance scalability to larger datasets and more complex architectures.

Second, our framework assumes access to sufficient data diversity within the unknown class cluster during training to generalize well to truly unseen categories. In practical scenarios, especially with long-tailed distributions, the representation of unknowns may be sparse or biased. Incorporating techniques such as data augmentation, synthetic unknown generation, or adversarial learning could help mitigate this issue.

Third, our method currently focuses on the feature-level information content without explicitly modeling semantic relationships among classes. Integrating semantic priors or hierarchical structures into the learning process may provide a more principled way to handle open-set categories.

In future work, we also plan to extend our approach to dynamic or continual learning settings, where the set of unknown classes evolves over time. This requires mechanisms for adapting the feature representation and updating the CMI objective without catastrophic forgetting. Exploring these directions would enhance the robustness and applicability of our method in real-world deployment scenarios.

\section{Concluding Remarks}

We propose Conditional Channel Capacity Maximization (3CM) as a general, plug-and-play regularization method for open-set semantic segmentation in 3D point clouds. Our experiments show consistent improvements across all benchmarks, achieving state-of-the-art performance without modifying the base architecture.

3CM stands out for its simplicity and generality. It can be integrated into any segmentation pipeline by augmenting the loss with an information-theoretic term that encourages models to retain more informative feature representations. This allows better discrimination of unknown classes while preserving performance on known categories.

Unlike prior works that rely on uncertainty modeling, clustering, or synthetic unknowns, our method provides a theoretically grounded approach based on mutual information. This not only improves robustness but also enhances interpretability of the learned features.

Although effective, 3CM assumes the availability of pseudo-unknown labels during training, which may not always hold in practice. Additionally, the computation of class-wise feature distributions introduces slight training overhead. Future work may explore more efficient approximations or combine 3CM with spatial priors for improved consistency in complex scenes.

{
    \small
    \bibliographystyle{ieeenat_fullname}
    \bibliography{main}
}


\end{document}